\documentclass[preprint,12pt]{elsarticle}

\usepackage{amssymb}
\usepackage{lineno,hyperref}
\usepackage{mathrsfs}
\usepackage{graphicx}
\usepackage{tabularx}
\usepackage{amsmath}
\usepackage{color} 
\usepackage{stfloats}
\usepackage{multirow}
\usepackage{booktabs}
\usepackage{caption}
\usepackage{subcaption}
\usepackage{pdflscape}
\modulolinenumbers[5]

\journal{Image and Vision Computing}

\begin{document}

\begin{frontmatter}

%% Title, authors and addresses

%% use the tnoteref command within \title for footnotes;
%% use the tnotetext command for theassociated footnote;
%% use the fnref command within \author or \address for footnotes;
%% use the fntext command for theassociated footnote;
%% use the corref command within \author for corresponding author footnotes;
%% use the cortext command for theassociated footnote;
%% use the ead command for the email address,
%% and the form \ead[url] for the home page:
%% \title{Title\tnoteref{label1}}
%% \tnotetext[label1]{}
%% \author{Name\corref{cor1}\fnref{label2}}
%% \ead{email address}
%% \ead[url]{home page}
%% \fntext[label2]{}
%% \cortext[cor1]{}
%% \affiliation{organization={},
%%             addressline={},
%%             city={},
%%             postcode={},
%%             state={},
%%             country={}}
%% \fntext[label3]{}

\title{Parallax-Tolerant Image Stitching with Epipolar Displacement Field}

%% use optional labels to link authors explicitly to addresses:
%% \author[label1,label2]{}
%% \affiliation[label1]{organization={},
%%             addressline={},
%%             city={},
%%             postcode={},
%%             state={},
%%             country={}}
%%
%% \affiliation[label2]{organization={},
%%             addressline={},
%%             city={},
%%             postcode={},
%%             state={},
%%             country={}}

\author[inst1,inst2]{Jian Yu}
\author[inst1,inst2]{Feipeng Da\corref{label3}}
\cortext[label3]{Corresponding author}
\ead{dafp@seu.edu.cn}
\affiliation[inst1]{organization={School of Automation, Southeast University},%Department and Organization
            addressline={No.2 Sipailou}, 
            city={Nanjing},
            postcode={210096}, 
            state={Jiangsu},
            country={China}}
\affiliation[inst2]{organization={Key Laboratory of Measurement and Control of Complex Systems of Engineering, Ministry of Education},%Department and Organization
            addressline={No.2 Sipailou}, 
            city={Nanjing},
            postcode={210096}, 
            state={Jiangsu},
            country={China}}

\begin{abstract}
%% Text of abstract
Image stitching with parallax is still a challenging task. Existing methods often struggle to maintain both the local and global structures of the image while reducing alignment artifacts and warping distortions. In this paper, we propose a novel approach that utilizes epipolar geometry to establish a warping technique based on the epipolar displacement field. Initially, the warping rule for pixels in the epipolar geometry is established through the infinite homography. Subsequently, the epipolar displacement field, which represents the sliding distance of the warped pixel along the epipolar line, is formulated by thin-plate splines based on the principle of local elastic deformation. The stitching result can be generated by inversely warping the pixels according to the epipolar displacement field. This method incorporates the epipolar constraints in the warping rule, which ensures high-quality alignment and maintains the projectivity of the panorama. Qualitative and quantitative comparative experiments demonstrate the competitiveness of the proposed method for stitching images with large parallax.
\end{abstract}

%%Graphical abstract
% \begin{graphicalabstract}
% \includegraphics{grabs}
% \end{graphicalabstract}

% %Research highlights
% \begin{highlights}
%     \item Novel warping rule based on the projective model: The algorithm proposes a unique warping rule that defines pixel warping in epipolar geometry using the infinite homography, ensuring projection consistency and a unified coordinate system in image stitching.
%     \item Thin-plate spline deformation for epipolar displacement field: The method leverages thin-plate splines to formulate the epipolar displacement field, enabling precise pixel displacement along the epipolar line and maintaining high-quality alignment in images with large parallax.
%     \item Region Expansion with Warping Rule: This method extends the alignment criterion beyond overlapping regions by incorporating epipolar constraints into the warping rule. It forces the points in non-overlapping regions to lie on the corresponding epipolar lines, ensuring the global projectivity of the panoramic image.
% \end{highlights}

\begin{keyword}
%% keywords here, in the form: keyword \sep keyword
image stitching \sep epipolar geometry \sep the infinite homography
%% PACS codes here, in the form: \PACS code \sep code
% \PACS 0000 \sep 1111
%% MSC codes here, in the form: \MSC code \sep code
%% or \MSC[2008] code \sep code (2000 is the default)
% \MSC 0000 \sep 1111
\end{keyword}

\end{frontmatter}

%% \linenumbers

%% main text
\section{Introduction}
\label{sec:intro}
Image stitching is a potent technique that has made notable advancements in diverse fields, including autonomous driving, medical imaging, surveillance video, and virtual reality \cite{wang2020review,pandey2019image,cao2020constructing}. It involves combining multiple images with a limited field of view to create a scene with a wider field of view. Despite the significant progress made in image stitching techniques over the past few decades, generating high-quality panoramic images remains a challenge, particularly when dealing with images with significant parallaxes.

Image stitching commonly employs a 3$\times$3 homography matrix, which represents a 2D projection transform, for image alignment. However, real-world scenes are often non-planar, or the viewpoints are not co-located, rendering a single global homography projection model inadequate in describing the required transformations. As a result, image misalignment or ghosting effects may occur. To mitigate parallax artifacts, representative existing methods include adaptive warping algorithms and shape preservation methods. Adaptive warping algorithms segment the image and employ distinct warping models \cite{gao2011constructing,zaragoza2013as,zhang2016multi,li2018parallax,Lee_2020_CVPR} for different regions to optimize the warping process using an energy minimization framework, thus reducing parallax artifacts \cite{zhang2016multi,li2018parallax}. Nevertheless, the use of multiple homography transformations may introduce inconsistencies among the perspective transformations, which can impact the natural appearance of the overall stitched image. Shape preservation methods aim to maintain both local and global geometric formations by leveraging geometric features, leading to improved stitching outcomes. Prominent geometric features, including frequently used feature points and line segments that retain the linear structure of an image, form substantial constraints for homography estimation when combined in image stitching  \cite{jia2021leveraging,li2015dual,liao2019single,xiang2018image}. Additionally, more intricate geometric features, such as edge contours \cite{du2022geometric}, depth maps \cite{liao2022natural} and semantic plane regions \cite{li2021image}, are employed when designing diverse energy functions to enhance content alignment and shape preservation. Nonetheless, the suitability of these intricate designs in real-world applications necessitates considering factors like the availability of adequate geometric support for the scene and the computational efficiency of the algorithm. Recently, techniques utilizing Convolutional Neural Networks (CNNs) for accurate homography estimation and stitching have emerged gradually. These methods discard geometric features in favor of high-level semantic features that can be flexibly learned using supervised \cite{song2021end,nie2020view,nie2022learning}, weakly supervised \cite{song2022weakly}, or unsupervised \cite{nie2021unsupervised} approaches. Although these methods deliver robust performance in stitching images with small baselines, they encounter difficulties when handling substantial parallax and conditions involving different datasets and resolutions.

Upon examining the aforementioned methods, it is evident that they are no longer reliant on fixed projection models. Instead, they adapt the models based on image data or salient geometric features in a data-driven manner, aiming to accurately align with the data.  While these approaches effectively remove artifacts, they may also lead to potential violations of projection principles in image stitching, resulting in the inclusion of undesired points in the stitched image due to foreground occlusion. This issue is more likely to occur in stitching scenarios with larger baselines. It is important to emphasize that the views utilized for stitching still adhere to the rules of projection transformation and are subjected to strict epipolar geometry constraints. Therefore, this paper proposes a parallax-tolerant stitching method based on the epipolar displacement field (EDF). The primary motivation of this paper is to maintain the global projection of panoramic images, which is equivalent to ensuring a unified coordinate system. This enables subsequent tasks such as object detection and pose estimation to be carried out on the stitched images. Consequently, a novel warping rule is devised based on the projection model, restricting the projection of stitched images to a singular viewpoint. Starting with the definition of planar-induced homography, we deploy the homography-based warping model in the epipolar geometry of two views, establishing a warping method based on the infinite homography and epipolar geometry. This method establishes a mapping geometric relationship for image stitching in the epipolar geometry, correcting the projection relationship of the stitched images to satisfy the epipolar geometry constraint. Then, we draw on the idea of elastic local alignment (ELA) \cite{li2018parallax} and extend the local elastic deformation to the epipolar displacement field for image warping. The displacement field is created using the mathematical model of thin-plate splines (TPS) \cite{rohr1996point}, whose parameters can be solved linearly according to the related theory of TPS. We use the model parameters to calculate the displacement of grid anchors in the image and generate the displacement field of the warped image through linear interpolation. Finally, the warped image is directly reprojected to generate seamless panoramic images. Overall, the algorithm retains the advantages of local deformation methods in accurately aligning image overlap regions and introduces the epipolar constraint into the displacement field. It applies this constraint in the propagation process of the warping rule from the overlap to non-overlap regions, achieving projection consistency of panoramic images while eliminating disparity artifacts. This also preserves the shape to some extent, avoiding the computational costs associated with extracting lines or other complex features. The experimental results demonstrate that the proposed algorithm accurately aligns images with large parallax and exhibits advantages in both qualitative and quantitative comparisons. To sum up, we conclude our contributions as follows:

\begin{itemize}
    \item A novel warping rule is formulated on the basis of the projective model, defining the pixel warping in epipolar geometry through the infinite homography. 
    \item Thin-plate spline deformation is utilized to construct the epipolar displacement field, indicating the pixel's displacement along the epipolar line. 
    \item By integrating epipolar constraints into the warping rules, this approach guarantees high-quality alignment and preserves the projectivity of the panoramic image.
\end{itemize}

The paper is organized as follows. In Section \ref{sec:related_work}, the relevant literature on image stitching is discussed. Section \ref{sec:method} provides a detailed description of the proposed method, which includes deriving the plane-induced homography in epipolar geometry model, estimating the infinite homography, establishing the epipolar displacement field, and acquiring panoramic images. The results are presented and compared with other methods in Section \ref{sec:experiment}. Section \ref{sec:conclusion} concludes the paper.

%% The Appendices part is started with the command \appendix;
%% appendix sections are then done as normal sections
% \appendix

\section{Related Work}
\label{sec:related_work}
Extensive research has been conducted on the fundamental principles of image stitching in the domains of computer vision and graphics. In this section, a concise survey is provided on adaptive warping methods, shape preservation methods, and recent advancements in deep learning.

\subsection{Traditional Stitching}
\textbf{Adaptive Warping Methods} 

Utilizing multiple homography transformations for adaptive alignment proves to be an effective approach in mitigating parallax artifacts. Gao \cite{gao2011constructing} \textit{et al.} introduced the dual-homography warping (DHW), where two homography matrices correspond to the perspective plane and the ground plane, and the stitching model is expressed as a combination of these two homography transformations. Zaragoza \cite{zaragoza2013as} \textit{et al.} presented the As-projective-as-possible (APAP) algorithm, which partitions the image into grids and estimates a single homography for each grid to achieve alignment. Li \cite{li2018parallax} \textit{et al.} employed robust elastic warping to mitigate parallax errors and devised a Bayesian feature refinement model to dynamically eliminate erroneous local matching in the grid. Lee \cite{Lee_2020_CVPR} \textit{et al.} substituted the spatial distance-based weights in APAP with warping residual weights and partitioned the input image into superpixels for adaptive warping. Nonetheless, the outcomes of this method are somewhat reliant on the accurate estimation of the initial multiple homographies.

Moreover, seam-driven image stitching \cite{gao2013seam,zhang2014parallax,lin2016seagull} techniques endeavor to identify parallax-free local regions for precise alignment followed by the generation of high-quality seams for seamless stitching. Similar to adaptive warping methods, they strive to achieve superior alignment to eliminate parallax artifacts, but generally struggle to preserve the geometric structure information in the image.

\textbf{Shape Preservation Methods}
Shape preservation methods aim to maintain both local and global geometries, resulting in visually pleasing seamless transitions. On one hand, methods such as shape-preserving half-projection (SPHP) \cite{chang2014shape}, adaptive as-natural-as-possible (AANAP) warping \cite{lin2015adaptive}, and global similarity prior (GSP) models \cite{chen2016natural} strive to enhance shape preservation by utilizing either local homography transformations or global similarity transformations. On the other hand, the combination of multiple geometric features in joint alignment can impose stronger constraints on homography estimation. Examples include the dual-feature warping (DFW) \cite{li2015dual} based on point-line features, the single-perspective warping (SPW) \cite{liao2019single}, the line-guided local warping \cite{xiang2018image} with global similarity constraints, and the GES (GEometric Structure preserving)-GSP \cite{du2022geometric} which leverages edge features for a seamless transition between local alignment and shape preservation.

It should be noted that the aforementioned traditional methods lack a consistent projection model, and the resulting stitched images might contravene the principles of projection. Thus, the main purpose of this article is to guide image stitching in accordance with the rigorous principles of projection.

\subsection{Deep Learning Solution}
Deep image stitching originates from advancements in deep homography estimation. Since the initial proposal of deep homography estimation networks by DeTone \cite{detone2016deep} \textit{et al.}, various supervised \cite{shao2021localtrans,cao2022iterative}, semi-supervised \cite{jiang2023semi}, and unsupervised \cite{nguyen2018unsupervised,kharismawati2020cornet,ye2021motion,wang2024mask} methods have been continuously proposed. These methods have demonstrated advanced performance in homography estimation, especially in scenarios with small baselines. However, significant room for improvement remains in wide-baseline scenes, particularly in those with multiple planes \cite{jiang2023semi}. 

Some deep image stitching methods employ homography estimation networks to estimate single \cite{nie2020view,nie2022learning} or multiple homographies \cite{song2021end,nie2021depth} to warp images. Given the unavailability of ground truth stitching labels, unsupervised approaches are preferred \cite{nie2021unsupervised}. However, there is a pressing need for further progress in deep image stitching, as the current state of the art is hindered by the limitations of deep homography estimation networks. Some approaches have augmented deep homography with local alignment adjustments. For instance, Jia \textit{et al.} explored texture and geometric consistency constraints to achieve non-uniform pixel-level alignment in overlapping regions. They utilized the overlapped areas to ensure consistency of content and structure in non-overlapping regions, thus refining the alignment of the entire stitched image. In a similar vein, Nie \textit{et al.} simultaneously parameterized homography transformation and thin-plate spline transformation within the network. The former facilitated global linear transformation, while the latter enabled local non-linear deformations, allowing for precise image alignment, especially in regions with parallax.

\section{Methodology}
\label{sec:method}
In this section, we begin by deriving the plane-induced homography in the epipolar geometry. This derivation allows us to establish the image warping formula based on the infinite homography and epipolar geometry. Subsequently, we expound on the methodology for retrieving the infinite homography under certain conditions, specifically for uncalibrated stereo rigs. Lastly, we construct an epipolar displacement field based on the warping model in epipolar geometry and the theory of local elastic deformation. This field guides the target image to warp into the panorama, ensuring both accurate alignment and satisfaction of the epipolar constraints.

\subsection{Plane Induced Homography}
\begin{figure}[t]
	\centering
	\includegraphics[width=\linewidth]{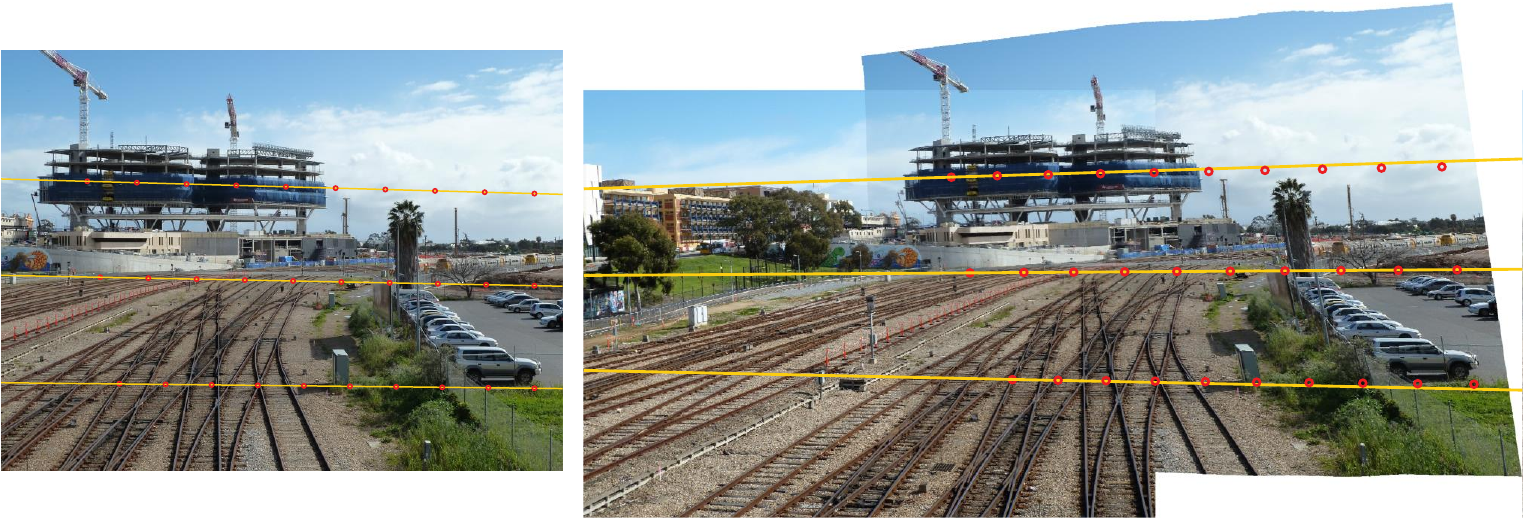}	
	\caption{Cases that fail to meet the epipolar constraint. Three epipolar lines, highlighted in yellow, along with several corresponding point sets, are identified in the upper half of the target image. However, the points that are transformed into the panoramic image using existing stitching methods no longer adhere to the epipolar lines as prescribed by the warping rule. This occurs mainly in the non-overlapping areas.}
	\label{fig:1}
\end{figure}

Image stitching utilizes planar homography to transfer points from one view to another view, as if they were projected onto one plane. Planar homography establishes the correspondence of image points between the two views, which is a form of projection relation closely associated with the two views' epipolar relationship. However, in many existing stitching methods, the use of multiple homographies for high-precision local alignment often results in projection distortion in non-overlapping regions, which hinders achieving better stitching quality. Alternatively, the adoption of a global similarity transformation method can also disrupt the epipolar relationship of particular points in non-overlapping regions. As demonstrated in Figure \ref{fig:1}, we labeled three epipolar lines and several points on each line in the target image for stitching. After applying existing stitching methods (\textit{e.g.} APAP), we relabel these points and epipolar lines in the stitched image following the warping rules. In the overlapping region, the warped points still satisfy the epipolar relationship, while the points in the non-overlapping region gradually deviate from the epipolar lines. Therefore, based on the plane-induced definition of homography, we propose a stitching model grounded in epipolar geometry that emphasizes the epipolar constraints in the stitched image.

Given a calibrated stereo rig, where the images in the two viewpoints are $\mathcal{I}$ and $\mathcal{J}$. Suppose the camera matrices of the two views are with the world origin at the first camera
\begin{equation}
\textbf{P} = \textbf{K}[\textbf{I}|\textbf{0}]\quad\text{and} \quad \textbf{P}' = \textbf{K}'[\textbf{R}|\textbf{t}]
\end{equation}
where $\textbf{K},\textbf{K}'\in\mathbb{R}^{3 \times 3}$ are intrinsics (camera calibration matrix) of the two cameras. $\textbf{R}\in \text{SO}(3)$ and $\textbf{t}\in\mathbb{R}^3$ are the rotational matrix and translation vector from the second camera's coordinate frame to the world coordinate frames.

Suppose a plane $\pi_E = (\textbf{n}^T, d)^T$ in the world frame which does not contain either of the camera centers. $\textbf{n}$ is the normal vector of the plane and $d/\|\textbf{n}\|$ is the orthogonal distance of the plane from the first camera. Let $\textbf{X}$ be a world point on $\pi_E$ so that $\textbf{n}^T\tilde{\textbf{X}}+d=0$. $\tilde{\textbf{X}}$ represents the inhomogeneous coordinate of $\textbf{X}$. The homography induced by $\pi_E$ can be expressed as
\begin{equation}
\textbf{H}=\textbf{K}'(\textbf{R} - \textbf{t}\textbf{n}^T/d)\textbf{K}^{-1}
\label{eq:2}
\end{equation}

Since the epipole $\textbf{e}'$ on $\mathcal{J}$ is $\textbf{K}'\textbf{t}$ and replacing $\textbf{m}$ with $-(\textbf{n}^T/d)\textbf{K}^{-1}$, the homography in Eq.(\ref{eq:2}) can be further expanded as
\begin{equation}
\textbf{H}=\textbf{K}'\textbf{R}\textbf{K}^{-1} + \textbf{e}'\textbf{m}^T = \textbf{H}_{\infty} + \textbf{e}'\textbf{m}^T
\label{eq:3}
\end{equation}
where $\textbf{H}_{\infty}=\textbf{K}'\textbf{R}\textbf{K}^{-1}$ is the infinite homography. Hence, the homography generated by a plane in space is a three-parameter family of homographies that are parameterized by $\textbf{m}$. It is defined by the plane, camera intrinsic parameters, and relative extrinsic parameters. In addition, $\textbf{m}$ also represents the vanishing line of the $\pi_E$ in $\mathcal{I}$.

\subsection{Acquiring the Infinite Homography}
To establish homography as defined by Eq.(\ref{eq:3}), we first need to obtain the infinite homography. If the aforementioned stereo rig is calibrated, such as in scenarios involving autonomous driving or security monitoring, the calibrated intrinsic and extrinsic parameters can be directly used to construct planar homography. However, for most image stitching cases, the rig is uncalibrated, which means that parameters such as the camera calibration matrix and rotation matrix are unknown.

Regarding the camera's intrinsic parameters, certain assumptions can be made, such as specifying an initial camera calibration matrix $\textbf{K}$ whose principal point coordinates can be approximated using the image's center point and assuming equal horizontal and vertical focal lengths in terms of pixel dimensions. The focal lengths can be estimated by examining the sensor chip size and lens focal length indicated in the image file header or by directly estimating a reasonable value. In the case of the stereo rig's extrinsic parameters, given the initial $\textbf{K}$ assumption, the rotation matrix can be derived by decomposing the fundamental matrix $\textbf{F}$, typically calculated using RANSAC with matched feature points $\{\textbf{x}_i, \textbf{x}_i'\}$ in the images \cite{hartley2003multiple,torr2000mlesac}. Moreover, the coordinates of epipoles $\textbf{e},\textbf{e}'$ in both views can be computed from the fundamental matrix according to $\textbf{Fe}=0$ and $\textbf{F}^T\textbf{e}'=0$. Finally, an optimization algorithm is utilized to refine the parameters by minimizing the geometric distance between the images. Here, an objective function is formulated using a first-order approximation of the geometric error for optimization purposes \cite{hartley2003multiple}
\begin{equation}
\underset{\textbf{\textbf{K},\textbf{K}',\textbf{R}}}{ \operatorname{arg\,min}}\sum_{i}^{}\frac{(\textbf{x}_i^T\textbf{H}_\infty\textbf{F}\textbf{x}_i)^2+(\textbf{x}_i'^T\textbf{F}\textbf{H}_\infty^{-1}\textbf{x}_i')^2}{(\textbf{F}\textbf{x}_i)_1^2+(\textbf{F}\textbf{x}_i)_2^2+(\textbf{F}'\textbf{x}_i')_1^2+(\textbf{F}'\textbf{x}_i')_1^2}
\end{equation}
where $(\textbf{F}\textbf{x}_i)_j^2$ represents the square of the $j$-th entry of the vector $\textbf{F}\textbf{x}_i$. It is important to note that optimized solutions for $\textbf{K},\textbf{K}'$ and $\textbf{R}$ are not unique, as different initial values for $\textbf{K},\textbf{K}'$ generate varying results. This is because the mapping from the camera matrices to the fundamental matrix is not injective (one-to-one). While the fundamental matrix obtained from the feature point pairs relates to a series of possible camera matrices, the computation of $\textbf{H}_\infty$ is unique due to the requirement of satisfying the epipolar geometry constraint of $\textbf{H}_\infty^T\textbf{F}=[\textbf{e}]_x$. \footnote{$[*]_x$ returns the skew-symmetric matrix of the input vector.}

\subsection{Warping via Epipolar Displacement Field}
\begin{figure}[h]
	\centering
	\includegraphics[width=\linewidth]{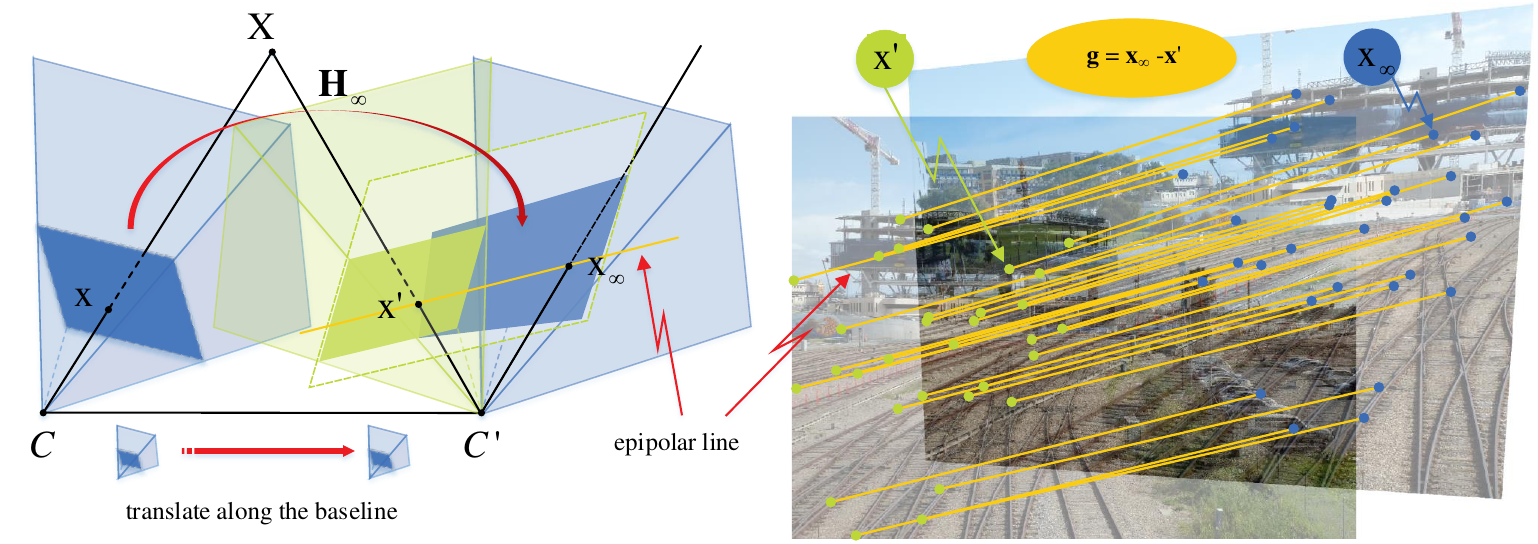}	
	\caption{Warp image with the Epipolar Displacement Field. The transformation based on the infinite homography is equivalent to shifting the viewpoint of the object along the baseline direction of the stereo rig towards the reference viewpoint, forming a concentric projection model. In the image plane of the reference viewpoint, the warped image is formed by the intersection of the field of view from the new viewpoint and the reference image plane. In the stitching viewpoint, both $\textbf{x}_\infty$ and $\textbf{x}'$ lie on the epipolar line to maintain the epipolar constraint, while their distances give rise to the epipolar displacement field. With the help of this field, image warping aligns the pixels to the exact position along the epipolar lines.}
	\label{fig:2}
\end{figure}

Similarly to warping with the global homography, we employ the infinite homography to map $\mathcal{I}$ onto the image plane of $\mathcal{J}$. As shown in Figure\ref{fig:2}, the corresponding points of $\textbf{X}$ in $\mathcal{I}$ and $\mathcal{J}$ are denoted as $\textbf{x}$ and $\textbf{x}'$, respectively. When $\textbf{X}$ is moved to infinity along its projection direction in $I$, the projection of $\textbf{X}$ in $\mathcal{J}$ will be $\textbf{x}_\infty$ transformed by the infinite homography. It can be seen that, in the epipolar plane where $\textbf{X}$ is located, $\textbf{x}$ in $\mathcal{I}$ and $\textbf{x}_\infty$ in $\mathcal{J}$ share the same direction of the counter-projection ray. Consequently, the transformation of $\mathcal{I}$ through the infinite homography can be understood as moving the optical center of $\mathcal{I}$ parallel along the baseline towards the optical center of the view $\mathcal{J}$ while maintaining the orientation and the size of the field of view (FOV) of $\mathcal{I}$ unchanged. The intersection of the new FOV of $\mathcal{I}$ at the new optical center with the plane of $\mathcal{J}$ yields the image obtained from $\mathcal{I}$ by the infinity homography. To further realize the stitching of the two views, it is necessary to continue to adjust $\textbf{x}_\infty$ to $\textbf{x}'$ along the epipolar lines.

Based on the technique of robust elastic warping (REW), a thin-plate spline curve with a simple radial basis function (RBF) type is selected to warp the image in the case of given anchor points, providing good performance in terms of alignment quality and efficiency. REW assigns the residuals between the corresponding points $\textbf{x}'$ and $\textbf{x}_\textbf{H}=\textbf{Hx}$ as $\textbf{g}$ and aims to minimize these errors. $\textbf{H}$ denotes the global homography. Specifically, the residual $\textbf{g}(u,v) = (g_x(u,v),h_y(u,v))^T$ represents a deformation of the image, with $g_x(u,v)$ and $h_y(u,v)$ defining the deformations in the $u$ and $v$ directions, respectively, and being mutually independent. The alignment term $J_D$ and the smoothness term $J_S$ together form the energy function for the optimal warp, which are defined as
\begin{equation}
\begin{aligned}
&J_D = \sum_{i=1}^{N}(g_x(u,v)^2+h_y(u,v)^2)\\
&J_S = {\iint_{}^{}}_{(x,y)\in\Omega }\left |\triangledown^2\textbf{g} \right |^2dxdy
\end{aligned}
\end{equation}

The entire energy function uses the parameter $\lambda$ to balance the two terms
\begin{equation}
J_\lambda = J_D+\lambda J_S 
\label{eq:opt}
\end{equation}

According to the regularization theory of TPS approximation, the deformation function can be represented as TPS interpolation. Take the $u$ direction as an example, the optimal solution to minimize Eq.(\ref{eq:opt}) can be expressed as
\begin{equation}
	g_x(u,v)=\sum_{i=1}^{N}\omega_i \phi_i(\textbf{x})+\alpha_1u+\alpha_2v+\alpha_3
	\label{eq:tps}
\end{equation}
where $\phi_i(\textbf{x}) = \left|\textbf{x}-\textbf{x}_H\right|^2\ln \left|\textbf{x}-\textbf{x}_H\right|$ is the RBF, $\textbf{w} =(\omega_1 , \cdots\omega_n)^T$ and $\textbf{a}=(\alpha_1,\alpha_2,\alpha_3)^T$ are the coefficients.

Retrospectively examining Eq.(\ref{eq:3}), we can express $\textbf{x}'$ as $\textbf{H}_\infty\textbf{x}+\textbf{e}'\textbf{m}^T\textbf{x}$, which, when represented in homogeneous coordinates, simplifies to $\tilde{\textbf{x}}'=\tilde{\textbf{x}}_\infty+\tilde{\textbf{e}}'$. Importantly, it is observed that $\textbf{m}$, a coefficient vector with three parameters, closely resembles the coefficient $\textbf{a}$ in Eq.(\ref{eq:tps}). Thus, we can formulate a displacement field between $\textbf{x}_\infty$ and $\textbf{x}'$ with reference to the form of the residual $\textbf{g}$. In this displacement field, all points are constrained to move along the corresponding epipolar lines. Therefore, we refer to it as an epipolar displacement field. These displacements can be expressed as
\begin{equation}
\begin{aligned}
&g_x(u,v)=\sum_{i=1}^{N}\omega_i \phi_i(\textbf{x}_\infty)+e_1'(m_1u+m_2v+m_3)\\
&h_y(u,v)=\sum_{i=1}^{N}\varpi_i \phi_i(\textbf{x}_\infty)+e_2'(m_1u+m_2v+m_3)
\end{aligned}
\label{eq:edf}
\end{equation}
where $\textbf{w}=[\omega_1,\cdots,\omega_N]^T,\textbf{w}'=[\varpi_1,\cdots,\varpi_N]^T$ are the RBF's coefficients for the two directions of the image. $e_1',e_2'$ are the two elements of the inhomogeneous coordinates of the epipole $\textbf{e}'$. These coefficient vectors can be computed through the following system of linear equations
\begin{equation}
\begin{aligned}
(\mathcal{K}+\rho\textbf{I})\textbf{w}+e_1'\textbf{M}\textbf{m}&=[g_{x1}(u,v),\cdots,g_{xN}(u,v))]^T\\
\textbf{M}^T\textbf{w}&=0\\
(\mathcal{K}+\rho\textbf{I})\textbf{w}'+e_2'\textbf{M}\textbf{m}&=[h_{y1}(u,v),\cdots,h_{yN}(u,v))]^T\\
\textbf{M}^T\textbf{w}'&=0
\end{aligned}
\end{equation}
where $\rho=8\pi$ is a constant to adjust the interpolation \cite{rohr1996point}, $\mathcal{K}$ contains the set of RBFs $\{\phi_i(\textbf{x}_\infty)\}$, and $\textbf{M}=[\textbf{x}'_i,\cdots,\textbf{x}'_N]^T$. 

\begin{figure}[ht]
	\centering		
	\includegraphics[width=0.7\linewidth]{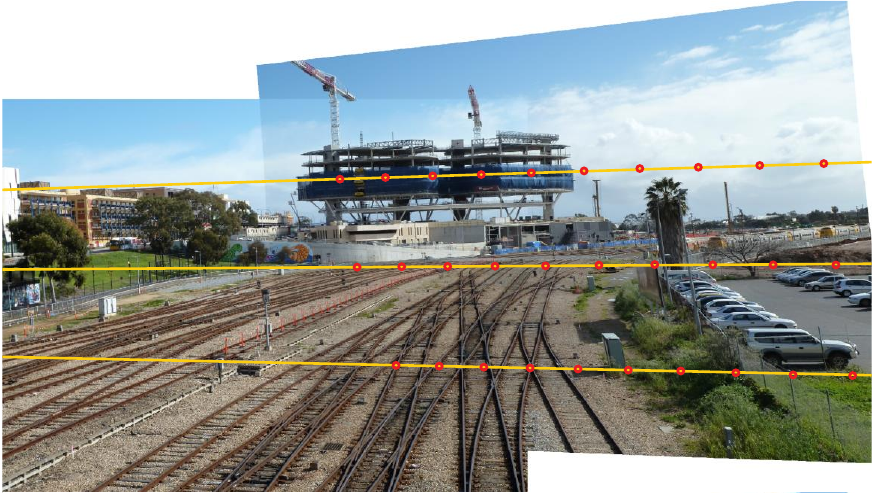}
	\caption{Epipolar constraints preservation: points on the epipolar line in the target image persist on the corresponding epipolar line after warping.}
	\label{fig:epi_preserve}
\end{figure}

Calculating the epipolar displacement field pixel-wise requires extensive computational resources. However, employing the approach outlined in APAP, we can partition the images into uniformly distributed grid cells and compute the displacement field of the cell vertices. Subsequently, interpolation can be utilized to ascertain the displacement field at every point within the cell. It should be noted that during the stitching process, the warping rules of the overlapping region often suffer from over-fitting when propagated to non-overlapping areas. Specifically, the locally smooth warping can disrupt the global projection in the non-overlapping area. The REW method mitigates this tendency by smoothly transforming the local deformations to the estimated global transformation, in particular the global homography. In this paper, we follow the smooth transition to global homography in their method, which is easily achieved by gradually reducing the deformation function $\textbf{g}(u,v)$ to zero in the non-overlapping region.Nevertheless, in our approach, we did not incorporate the warping model with the global similarity transformation, as done in the REW method. The proposed epipolar displacement field, introduced in this paper, suppresses the deformation of the non-overlapping area through the application of global infinite homography and epipolar constraints. Figure\ref{fig:epi_preserve} demonstrates that the offset of each point from the epipolar line has been amended compared to the results obtained in Figure\ref{fig:1}.

\section{Experiment}
\label{sec:experiment}
In this section, we conduct a qualitative and quantitative comparison of the EDF method proposed in this article with existing methods to assess its stitching quality performance. The comparison includes adaptive warping with multiple homographies (APAP \cite{zaragoza2013as}) , the elastic deformation method (REW \cite{li2018parallax}), single-projective warping (SPW \cite{liao2019single}), the line-point consistent shape preserving warping (LPC \cite{jia2021leveraging}) and the unsupervised deep image stitching (UDIS2 \cite{nie2023parallax}). The image pairs used in the experiments are primarily sourced from the literature of the aforementioned comparison methods.\footnote{https://github.com/dut-media-lab/Image-Stitching/tree/main/Imgs}$^,$\footnote{http://web.cecs.pdx.edu/~fliu/project/stitch/dataset.html} These image pairs are primarily captured from different viewpoints with large baselines, resulting in significant parallax. Additionally, they all possess adequate overlapping regions to assess the alignment performance of the stitching methods.

In the experiments, we employed VLFeat for extracting and matching SIFT features of the images, utilized RANSAC to eliminate outliers in the feature points, and applied the eight-point method to compute the fundamental matrix of the stereo rig. As outlined in Section \ref{sec:method}, our method initially estimated the camera's intrinsic parameters to establish an initial calibration matrix $\textbf{K}$. Subsequently, the infinite homography $\textbf{H}_\infty$ from the reference image to the target image was obtained through a geometric error optimization. Concerning the parameter configurations of the comparison methods, APAP set the $\sigma$ within the range of 8-12.5, and assigned $\gamma=0.01$. The weighted parameter $\lambda$ in REW was assigned a value of 0.1\% of the product of the length and width of the image, while the smooth transition width in non-overlapping regions was set to five times the maximum bias. In order to ensure fairness, our method used the same parameters as REW in the TPS deformation, and the cell size was fixed to 10 × 10 pixels for all methods requiring cell division in the comparison. In LPC, the threshold $\mu$ was established as 3 times the diagonal length of the grid. The $\lambda_{lo}$ and $\lambda_{gl}$ values in the energy function were set to 50 and 100, respectively. In addition, $\lambda_p$ and $\lambda_l$ were set to 1 and 5, and $\lambda_{dg}$ and $\lambda_{dn}$ are set to 50 and 100, correspondingly. SPW retains the identical parameters as LPC. To enable a comparison of alignment effects, all methods except UDIS2 employed linear blending for stitching. For UDIS2, we utilized the pre-trained model provided by the authors for warping and composition.

%-------------------------------------------------------------------------
\subsection{Qualitative comparison}
\begin{figure}[pt]
	\centering		
	\includegraphics[width=0.95\linewidth]{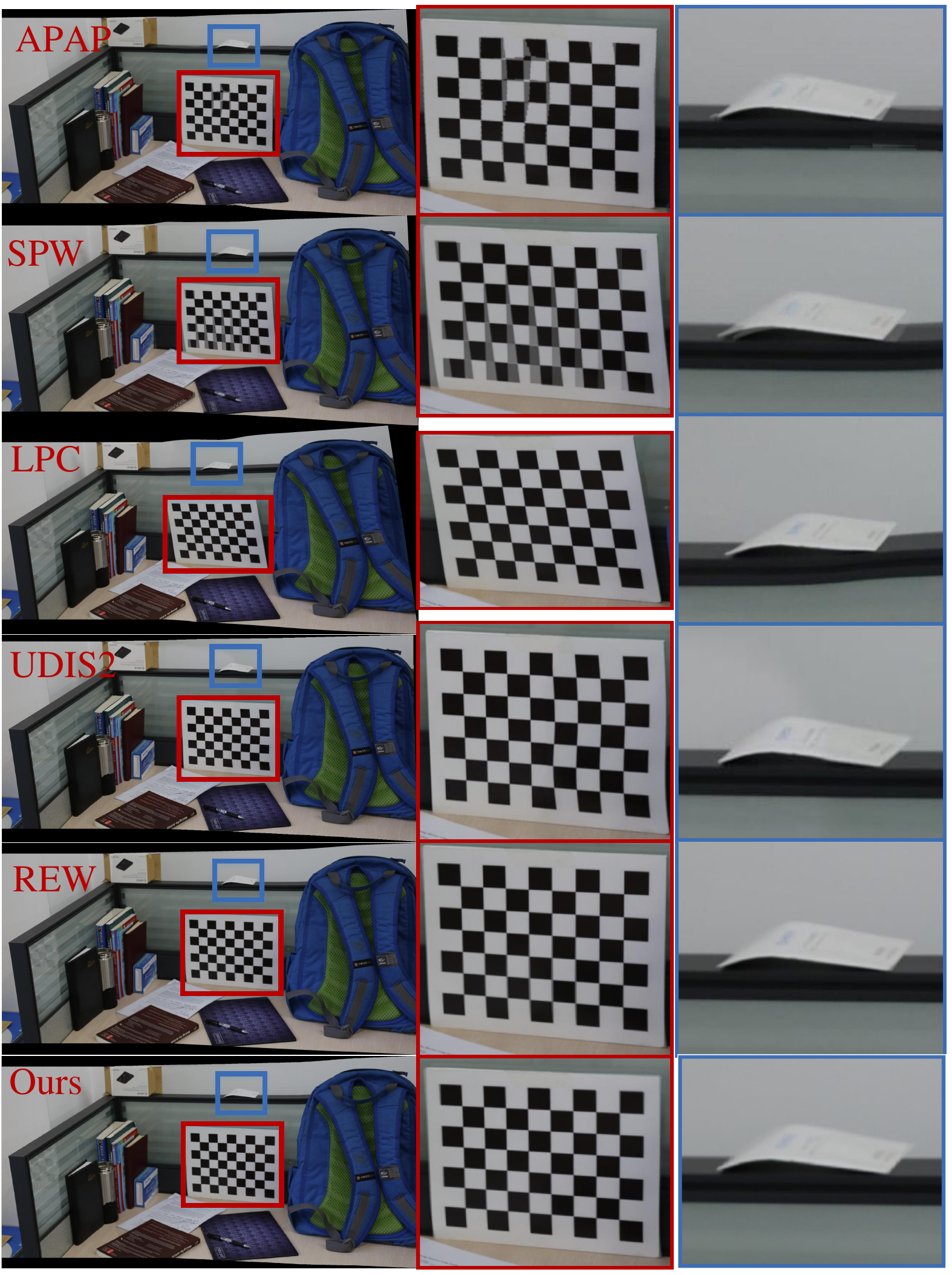}
	\caption{Comparison of stitching quality on a work table scene. From top to bottom are the results using APAP, SPW, LPC, UDIS2, REW and the method proposed in this paper. Regions are extracted and enlarged to show the alignment performance of the different methods in this challenging scenario.}
	\label{fig:worktable}
\end{figure}

Figure \ref{fig:worktable} illustrates the comparison of stitching results for a work table scene from the literature\cite{li2018parallax}, displaying the sequential presentation of results from top to bottom: APAP, SPW, LPC, UDIS2, REW and our method. The scene includes a calibration board with repetitive textures, and mismatches in these repetitive and structured patterns can cause pronounced distortion, thereby challenging the stitching algorithms. By zooming in on the chessboard section of the stitching results for each method, we observe multiple instances of misalignment and noticeable artifacts in both APAP and SPW. While LPC successfully aligns the target, it manifests bending deformation in the middle section on the top of the calibration board. This deformation extended to other areas of the image, leading to an overall geometric distortion. The same issue is observed in UDIS2. Our method and REW yield comparable outcomes by accurately aligning the repetitive regions of the chessboard. In regions like the white sticky note area with weak texture in the figure, noticeable artifacts were present in both APAP and SPW, whereas REW and our method exhibited superior artifact removal capabilities. Particularly noteworthy is the effectiveness of UDIS2 in addressing such weakly textured areas.

\begin{figure}[pt]
    \centering
    \subfloat[APAP-railtracks]{\includegraphics[width=0.9\linewidth]{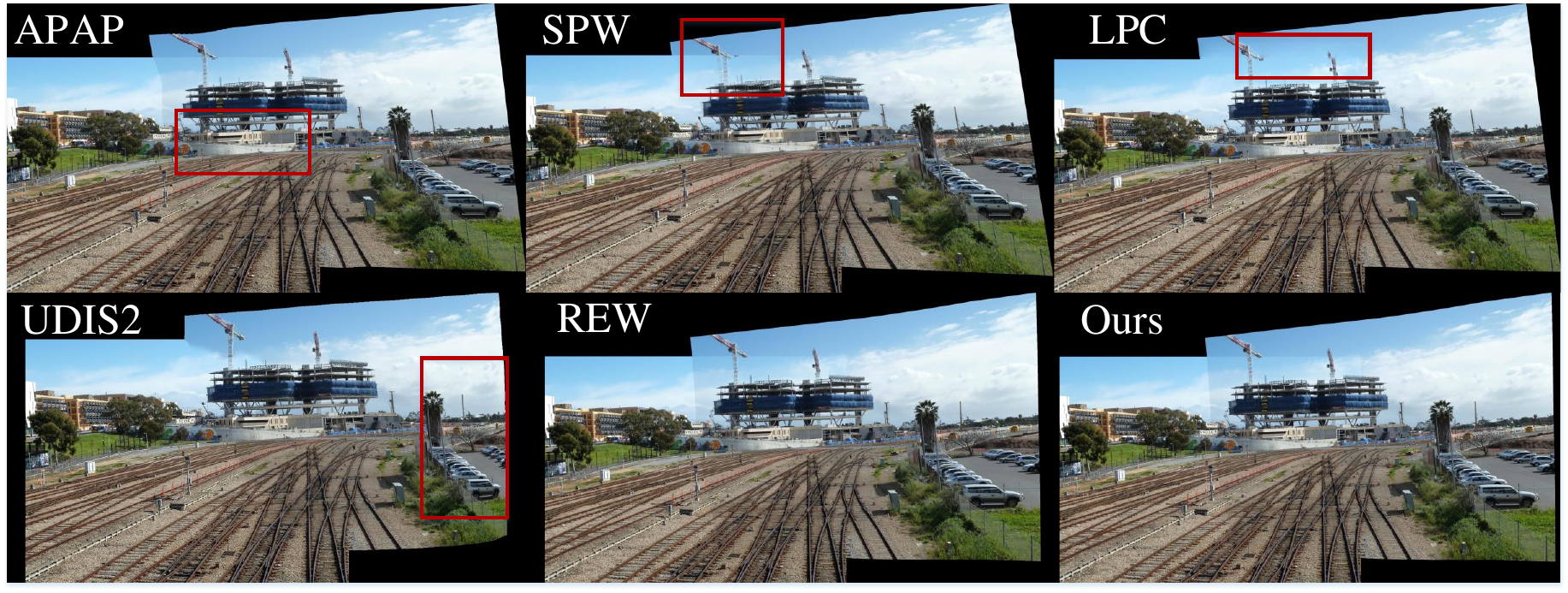}}
    \hfill            
    \subfloat[SPHP-garden]{\includegraphics[width=0.9\linewidth]{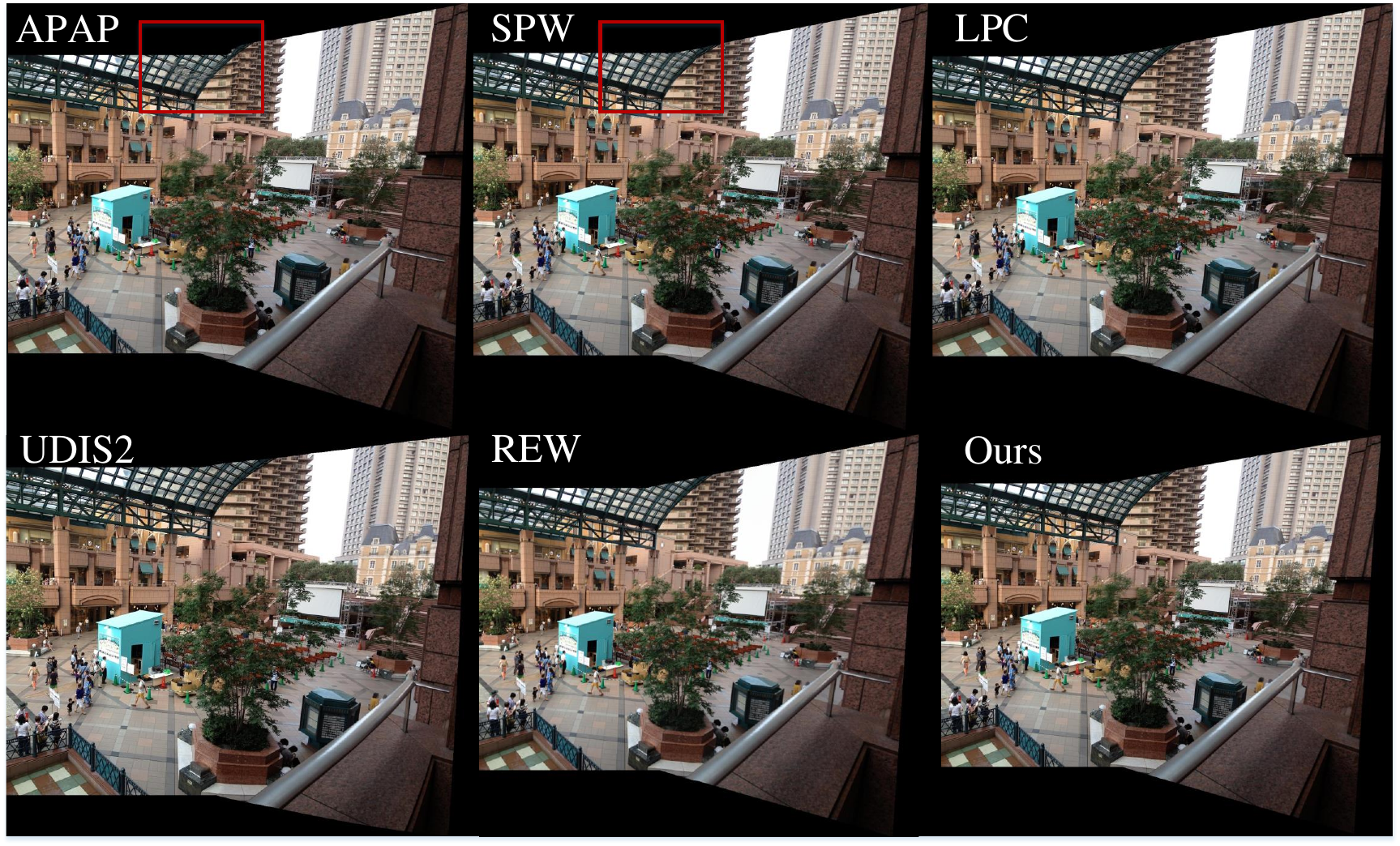}}
    \hfill
    \subfloat[PTS-063]{\includegraphics[width=0.9\linewidth]{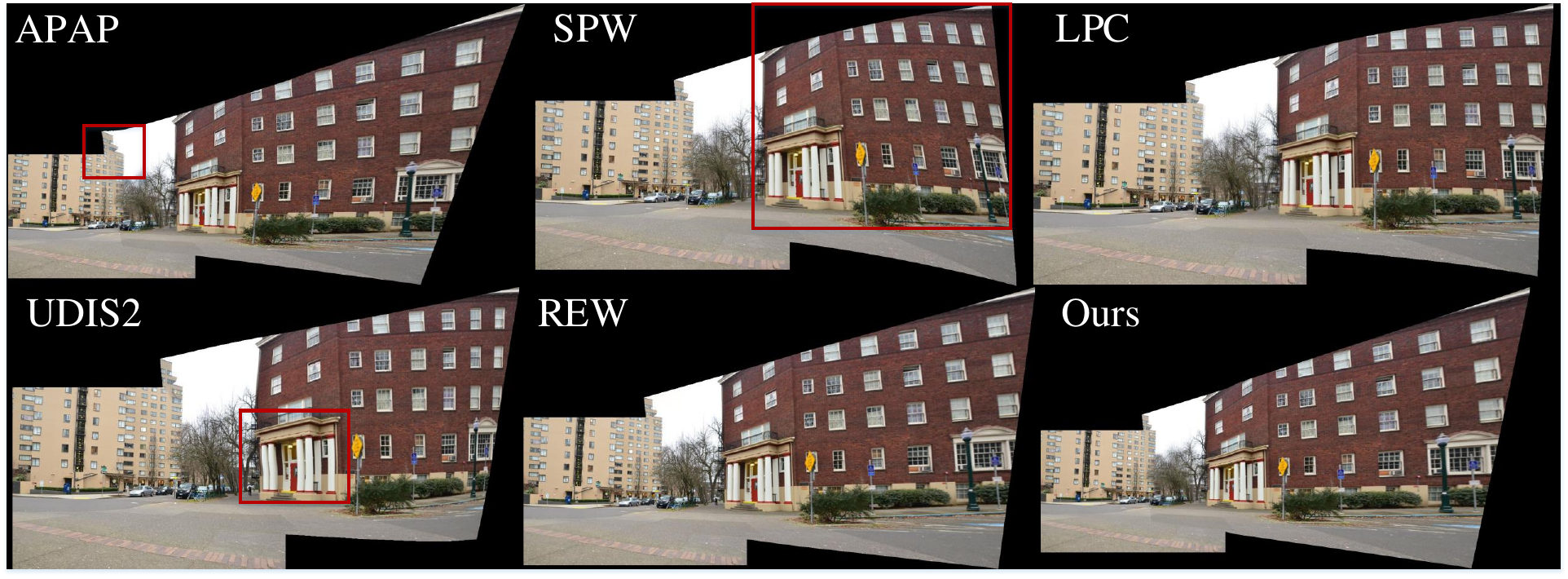}}
    \hfill       
    \caption{More Comparisons of the image stitching results obtained by the proposed method with APAP, SPW, LPC, UDIS2 and REW. The test image pairs are derived from the open source code of the literature \cite{jia2021leveraging,zhang2014parallax}.}
    \label{fig:short}
\end{figure}

Figure \ref{fig:short} displays additional scene comparisons includes railtracks \cite{zaragoza2013as}, garden \cite{chang2014shape} and neighborhoods (063) \cite{zhang2014parallax}. The figure marks regions with stitching imperfections. The railtracks serve as quintessential illustrations of extensive parallax stitching. Likewise, garden and neighborhoods exhibit analogous characteristics. Despite their modest baseline distance between viewpoints, a significant discrepancy in view direction prevails. In summary, the latter four methods yield results with fewer artifacts, superior stitching outcomes compared to APAP and SPW, and enhanced robustness. It is important to note that APAP employs a Gaussian function to assign weights to each mesh, resulting in fluctuations in certain meshes. For instance, fluctuations can be observed in the mesh that contains the tower crane in railtracks. SPW encounters local alignment issues in both the railtracks and garden scenarios, while the right building of its stitching results on neighborhoods deviate from global projectivity. Our forthcoming exploration will further elucidate the contrast in global projectivity through pertinent metrics in the quantitative experiments.

%-------------------------------------------------------------------------
\subsection{Quantitative comparison}
We employ Structural Similarity (SSIM) and Peak Signal-to-Noise Ratio (PSNR) as quantitative metrics to evaluate the registration quality of various methods. The tested images are obtained from publicly available data provided by references \cite{gao2011constructing,zaragoza2013as,li2018parallax,jia2021leveraging}. For maintaining fairness in the experiments, we employ the Multi-GS method for feature point matching across all methods to eliminate outliers, while not employing the Bayesian feature refinement model proposed in REW, which has been demonstrated to enhance alignment effects. Firstly, we detect and segment the overlapping region of the stitched images and calculate the SSIM and PSNR value for this region, which compares the reference and target images after stitching. The scores of APAP, SPW, LPC, REW and our method in 12 large parallax image experiments are presented in Table \ref{tab:ssim-psnr}, with the highest scores in each scenario highlighted in bold. Since the second stage of composition in UDIS2 employs seam stitching, it is not included in the comparison presented in Table \ref{tab:ssim-psnr}. The table illustrates that REW and our method achieve similar scores, both displaying the highest SSIM and PSNR values for numerous scenarios, indicating effective alignment effects in the overlapping regions for both approaches. Since both approaches employ TPS interpolation for constructing the deformation function, they demonstrate robust elastic alignment for local regions. Additionally, the deformation mechanism based on multiple control points mitigates the impact of computational errors in the epipole coordinates derived from the fundamental matrix.

\begin{table}[h]
	\centering
 \resizebox{\textwidth}{!}{
	\begin{tabular}{cccccc}
		\toprule
		\multirow{2}{*}{Images} & \multicolumn{5}{c}{SSMI/PSNR} \\
		& APAP & SPW & LPC & REW & Ours\\
		\midrule
		DHW-temple         & 0.658/27.186 & 0.522/24.214 & 0.892/26.428 & 0.934/29.657 & \textbf{0.943/30.240}\\	
		APAP-railtrack     & 0.619/23.715 & 0.451/21.904 & 0.882/21.959 & 0.926/26.721 & \textbf{0.929/27.102}\\	
		REW-boardingbridge & 0.615/29.300 & 0.585/28.560 & 0.854/27.950 & \textbf{0.897/30.881} & 0.886/30.713\\	
		SPHP-park          & 0.648/26.007 & 0.625/24.130 & \textbf{0.968}/24.822 & 0.911/\textbf{31.277} & 0.894/30.327 \\	
		DFW-desk           & 0.906/27.094 & 0.886/26.533 & 0.913/25.234 & 0.943/\textbf{30.418} & \textbf{0.979}/29.963 \\	
		REW-campussquare   & 0.706/29.638 & 0.525/28.742 & 0.937/27.454 & \textbf{0.955/34.447} & \textbf{0.955}/34.256 \\	
		SPHP-garden        & 0.565/26.115 & 0.551/26.092 & \textbf{0.952}/27.216 & 0.904/\textbf{29.130} & 0.894/28.737 \\		
		REW-cabin          & 0.728/25.242 & 0.523/21.885 & 0.874/22.407 & \textbf{0.962/30.120} & 0.950/29.062 \\	
		REW-gym            & 0.667/25.853 & 0.561/25.529 & 0.847/24.519 & 0.956/30.874 & \textbf{0.958/31.687} \\	
		REW-worktable      & 0.788/22.489 & 0.792/22.733 & 0.910/20.243 & \textbf{0.973}/28.911 & \textbf{0.973/28.994} \\	
        PTS-063            & 0.488/29.016 & 0.516/26.427 & 0.934/24.660 & \textbf{0.863}/29.935 & \textbf{0.863/30.327} \\	
        PTS-069            & 0.671/26.004 & 0.636/23.858 & 0.922/24.326 & \textbf{0.919/27.515} & 0.916/27.384 \\	
		\bottomrule
	\end{tabular}}
	\caption{Comparisons on SSIM and PSNR.}
	\label{tab:ssim-psnr}
\end{table}

%global projectivity
\begin{figure}[h]
    \centering
     \includegraphics[width=\linewidth]{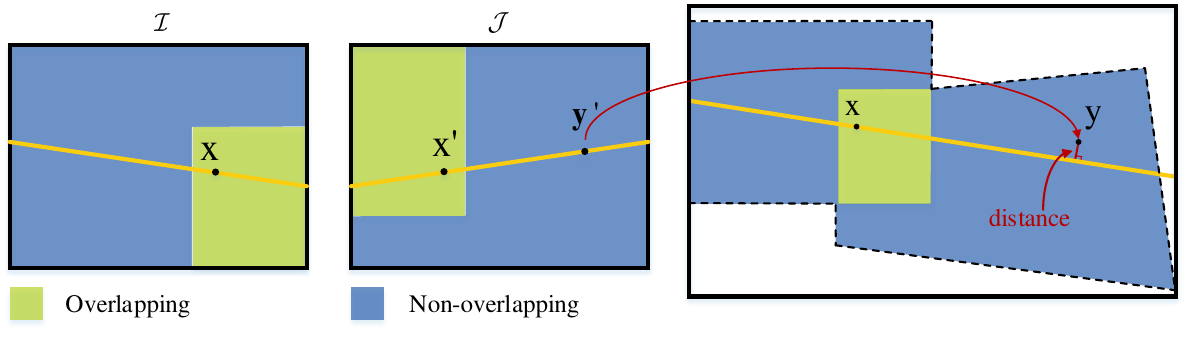}      
    \caption{Schematic of global projectivity quantization. Choose point $\textbf{y}'$ from the non-overlapping area of $\mathcal{J}$, then calculate the distance from its corresponding position $\textbf{y}$ in the stitched image to the associated epipolar line.}
    \label{fig:metric}
\end{figure}

\begin{table}[h]
	\centering
  \resizebox{0.6\textwidth}{!}{
	\begin{tabular}{cccc}
		\toprule
		Images & mathes & REW (px) & Ours (px)\\
		\midrule
            DHW-temple          & 179  & 8.747 & 7.216\\
		APAP-railtrack      & 1197 & 20.807 & 14.631 \\
            REW-boardingbridge  & 702  & 35.738 & 24.204\\
            SPHP-park           & 368  & 9.649 & 8.069\\
		DFW-desk            & 92   & 5.496 & 0.738 \\
		REW-campussquare    & 509  & 4.092 & 1.562\\
            SPHP-garden         & 675  & 16.161 & 10.228\\
            REW-cabin           & 765  & 6.139 & 2.176\\
            REW-gym             & 731  & 14.332 & 9.740\\
		REW-worktable       & 755  & 9.345 & 6.509\\
            PTS-063             & 332  & 6.750 & 5.792\\
            PTS-069             & 332  & 7.553 & 6.434\\
		\bottomrule
	\end{tabular}}
	\caption{Metric comparisons of global projectivity.}
	\label{tab:distance}
\end{table}

In addition to aligning overlapping regions, our approach also preserves the global projectivity of the stitched images, particularly by extending the epipolar constraint to non-overlapping areas. To demonstrate the global projectivity of our method, we introduce a comparison metric. Figure \ref{fig:metric} depicts two images, $\mathcal{I}$ and $\mathcal{J}$, showing their overlapping and non-overlapping regions. Matched feature points $\textbf{x}$ and $\textbf{x}$' lie on their respective epipolar lines. As all points on an epipolar line belong to the same epipolar plane, when points on the epipolar lines in $\mathcal{J}$ are mapped to $\mathcal{I}$, they remain on their corresponding epipolar lines. Therefore, within the non-overlapping area of $\mathcal{J}$, a point $\textbf{y}'$ is selected along the epipolar line of $\textbf{x}$', and the stitching algorithm identifies $\textbf{y}'$ as the corresponding point in the final stitched image denoted by $\textbf{y}$. The global projectivity is quantified by the distance from point $\textbf{y}$ to the corresponding epipolar line $\textbf{F}^T\textbf{x}'$; a smaller value indicates better preservation of global projectivity in the stitching process. 

\begin{figure}[t]
    \centering
    \subfloat[REW-cabin]{\includegraphics[width=\linewidth]{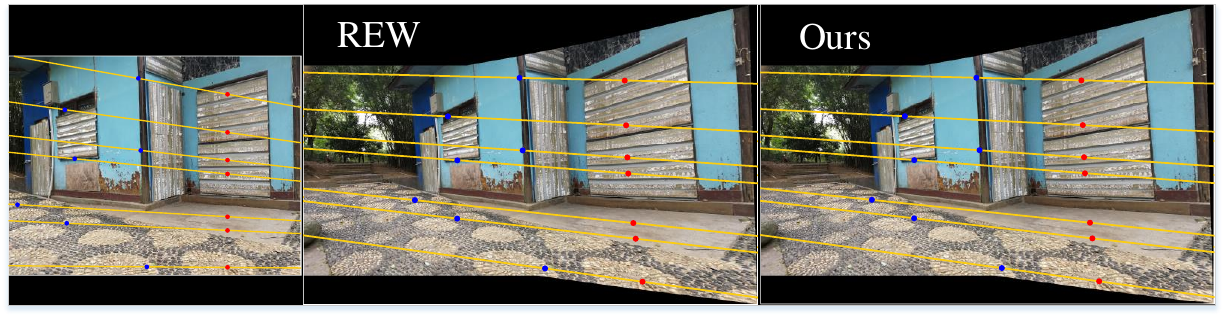}}
    \hfill            
    \subfloat[PTS-069]{\includegraphics[width=\linewidth]{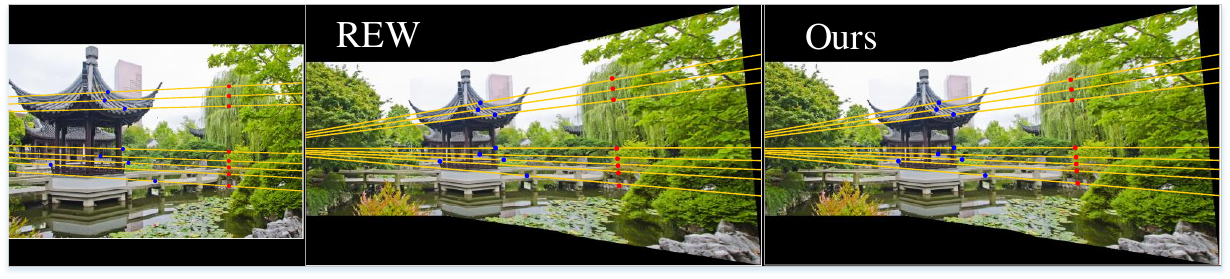}}
    \hfill      
    \caption{Example of global projectivity contrast.}
    \label{fig:projectivity}
\end{figure}

We used this metric to compare this paper's method with REW in several scenarios. For each scenario we selected $\textbf{y}'$ for each pair of matches and counted the mean values of the distances from all $\textbf{y}$ to the corresponding polar lines in Table \ref{tab:distance}. From the table, it can be seen that the distance value obtained by the our method is the smallest, which indicates that the epipolar constraint well limits the position of points in the non-overlapping regions in the stitching image and maintains a better global projectivity. We have chosen two scenes to demonstrate in Figure \ref{fig:projectivity}, with the leftmost image representing the J image intended for stitching. Feature points $\textbf{x}'$ have been highlighted in blue, points $\textbf{y}'$ in red, and their associated epipolar lines in yellow. Subsequently, we have annotated the mapped points $\textbf{y}$ on the images obtained through both REW and our method. It is noticeable that in our approach, the distance between $\textbf{y}$ and the corresponding epipolar line is reduced. This alignment ensures that points within the same epipolar plane in non-overlapping regions are confined to their respective epipolar lines, thereby expanding the warp rules from overlapping to non-overlapping areas.

%  computaional cost
Moreover, we assessed the runtime of various algorithms in the above scenario experiments. To reduce the variability introduced by steps such as feature extraction during stitching, Table \ref{tab:runtime} displays the average time from 100 experiment repetitions. All test methods were executed under the same experimental setup with a 3.6G-Hz CPU and 32GB RAM. The results demonstrate that our method has a significant advantage in computational efficiency, operating independently of scene texture information unlike the LPC and SPW methods. Specifically, in scenarios where scenes display distinct dual point-line attributes, these methods encounter extended processing times for feature extraction and matching.
\begin{table}[t]
	\centering
 \resizebox{\textwidth}{!}{
	\begin{tabular}{cccccccc}
		\toprule
		Images & image size (px) & APAP (s) & SPW (s) & LPC (s) & REW (s) & Ours (s)\\
		\midrule
		DHW-temple          & 730$\times$487   & 3.930  & 1.315 & 2.018  & 1.352 & 0.736\\		
		APAP-railtrack      & 1195 & 12.123 &104.497&1188.567& 10.733& 11.704\\
		REW-boardingbridge  & 2160$\times$1440 & 13.048 & 32.013&190.082 & 13.511& 12.920\\
		SPHP-park           & 1920$\times$1440 & 7.615  & 9.304 & 13.835 & 8.048 & 11.062\\
		DFW-desk            & 500$\times$375   & 3.554  & 0.616 & 0.624  & 0.997 & 0.370\\
		REW-campussquare    & 1280$\times$960  & 5.457  & 7.835 & 24.882 & 3.812 & 3.161\\
		SPHP-garden         & 1920$\times$1440 & 11.716 & 58.908&724.976 & 10.169& 10.143\\		
		REW-cabin           & 1280$\times$960  & 5.775  & 13.915& 70.572 & 3.855 & 3.524\\	
		REW-gym             & 1280$\times$960  & 5.697  & 51.064& 118.414& 3.481 & 3.085\\
		REW-worktable       & 2160$\times$1440 & 10.140 & 54.630& 48.151 & 8.421 & 8.058\\	
            PTS-063             & 800$\times$530   & 3.701  & 1.216 & 1.862  & 1.703 & 0.950\\	
            PTS-069             & 1000$\times$662  & 4.251  & 4.743 & 23.762 & 2.424 & 1.817\\	
		\bottomrule
	\end{tabular}}
	\caption{Comparisons on runtime.}
	\label{tab:runtime}
\end{table}
%-------------------------------------------------------------------------

\subsection{Limitations}
\begin{figure}[t]
	\centering
	\centering		
	\begin{subfigure}{0.6\linewidth}
		\includegraphics[width=\linewidth]{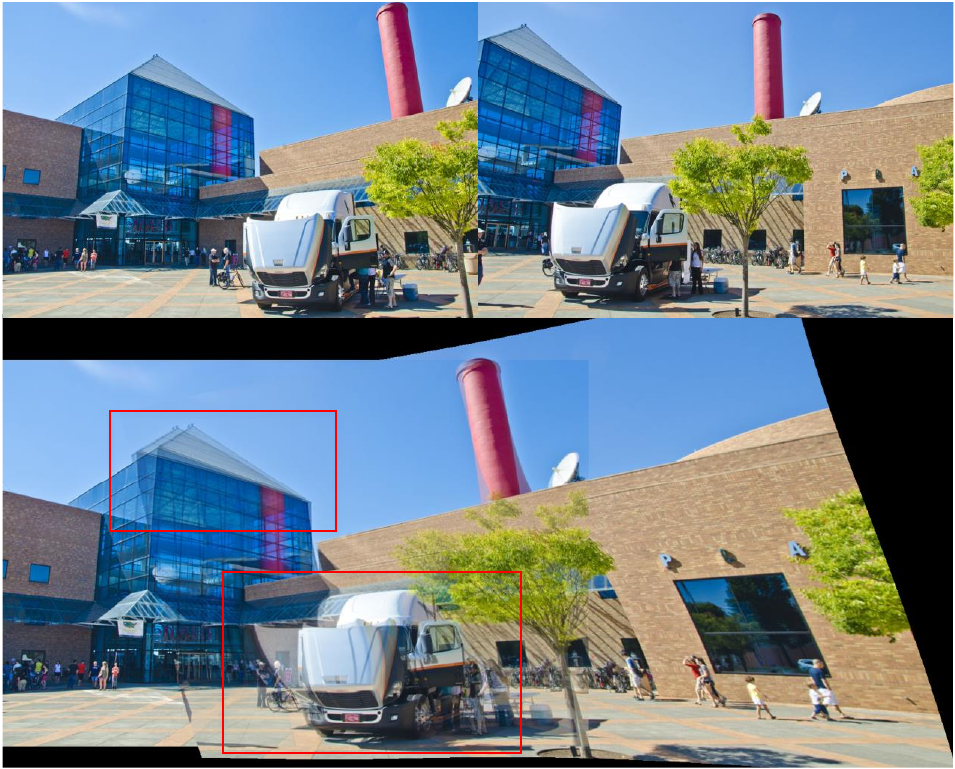}
		\caption{truck}
	\end{subfigure}
	\hfill
	\begin{subfigure}{0.6\linewidth}
		\includegraphics[width=\linewidth]{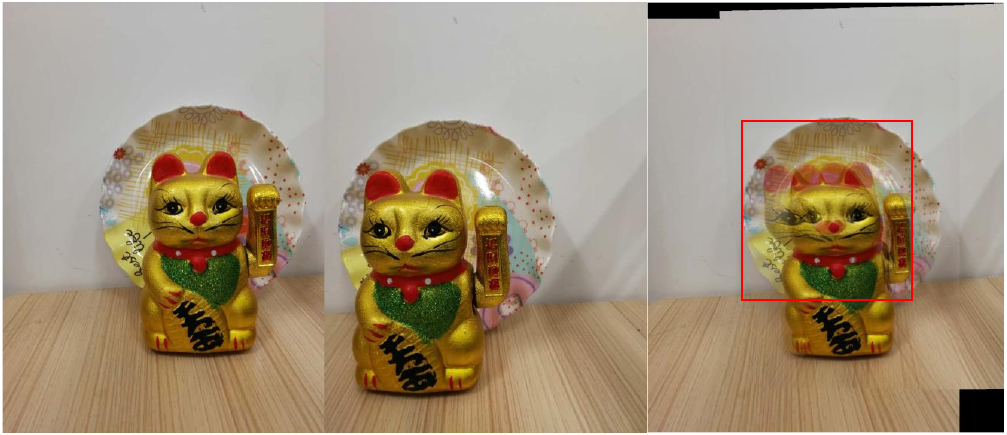}
		\caption{Maneki-neko}
	\end{subfigure}
	\caption{Failed cases were observed in scenarios due to severe occlusions in the input images. The identified artifacts in the resulting panoramas are highlighted with red circles.}
	\label{fig:fail_case}
\end{figure}

This method is primarily limited by the invisibility of spatial points from this particular perspective, which manifests in two different scenarios. First, when using a single perspective projection for stitching, the field of view larger than $180^\circ$ becomes theoretically invisible and cannot be effectively stitched together. This limits its ability to stitch multiple images. Second, there is the issue of invisibility caused by occlusion. Due to the implementation of a backward and reverse method, the inability to discern whether multiple points are projected onto the same location leads to an incapacity to detect occlusion. Figure \ref{fig:fail_case} illustrates several failed cases, where changes in the convergence of the perspective can readily lead to occlusions of the foreground objects between the two views, resulting in noticeable parallax artifacts. This convergence change mostly occurs when the perspectives' principal axes intersect forward.

\section{Acknowledgments}
This paper was partly supported by the National Natural Science Foundation of China (62305055), the Jiangsu Funding Program for Excellent Post-doctoral Talent (2022ZB118), and the Special Project on Basic Research of Frontier Leading Technology of Jiangsu Province of China (BK20192004C).

\section{Conclusion}
\label{sec:conclusion}
This paper proposes a novel parallax-tolerant image stitching approach with the epipolar displacement field. Our approach first utilizes the infinite homography to establish a warping technique based on knowledge of epipolar geometry. The infinite homography aligns the target perspective with the reference perspective to form a concentric projection model, which ensures the warping of the target image pixels is performed along the associated epipolar lines. Subsequently, drawing inspiration from elastic local alignment, we extend the concept of local elastic deformation to generate an epipolar displacement field, which is then formulated with thin-plate splines. To expedite the calculation of the epipolar displacement field without compromising alignment accuracy, we establish a uniform grid on the image plane and obtain the displacement field through interpolation of the grid anchor points. Finally, the warped image is transformed using the displacement field to produce the stitched result. Through qualitative and quantitative comparative experiments, we evaluate the performance of the proposed method, which has high alignment quality for overlapping regions and preserves shape in non-overlapping regions through epipolar constraints.

%% If you have bibdatabase file and want bibtex to generate the
%% bibitems, please use
%%
 \bibliographystyle{elsarticle-num} 
 \bibliography{cas-refs}

%% else use the following coding to input the bibitems directly in the
%% TeX file.

% \begin{thebibliography}{00}

% %% \bibitem{label}
% %% Text of bibliographic item

% \bibitem{}

% \end{thebibliography}
\end{document}